\documentclass[10pt,twocolumn,letterpaper]{article}

\usepackage[pagenumbers]{cvpr}

\usepackage{placeins}
\usepackage{cuted} %

\usepackage{currfile} %

\usepackage{caption} %

\usepackage{tikz}
\usetikzlibrary{positioning,arrows.meta,calc,fit,backgrounds}
\usepackage{bm}
\newcommand{\ahat}{\hat{\alpha}}
\newcommand{\acorr}{\hat{\alpha}'}
\newcommand{\Uband}{\mathcal{U}}
\newcommand{\Php}{\mathcal{P}_{\mathrm{hp}}}
\newcommand{\LAE}{\mathcal{L}_{\mathrm{AE}}}

\DeclareMathOperator{\clip}{clip}
\DeclareMathOperator{\mean}{mean}

\definecolor{cvprblue}{rgb}{0.21,0.49,0.74}
\usepackage[pagebackref,breaklinks,colorlinks,allcolors=cvprblue]{hyperref}

\def\confName{CVPR}
\def\confYear{2027}

\title{High-Fidelity Image Matting with Pixel-Space Diffusion Transformers}

\author{Baihan Yang\textsuperscript{1} \quad Xin Lin\textsuperscript{1} \quad Yufan Wei\textsuperscript{1} \quad Baiang Li\textsuperscript{2}\\
Yuheng Liu\textsuperscript{3} \quad Juanxi Tian\textsuperscript{4} \quad Xiaohui Xie\textsuperscript{3} \quad Truong Nguyen\textsuperscript{1}\\[2pt]
\textsuperscript{1}University of California, San Diego \quad \textsuperscript{2}Princeton University\\
\textsuperscript{3}University of California, Irvine \quad \textsuperscript{4}Nanyang Technological University
}

\begin{document}
\maketitle
\begin{strip}
\centering
\includegraphics[width=\textwidth]{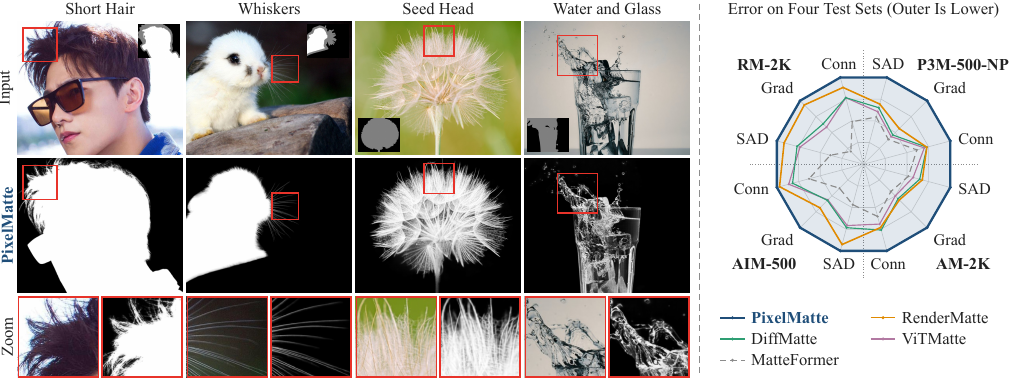}
\captionsetup{hypcap=false}
\captionof{figure}{PixelMatte performs generative trimap matting in one step, directly in pixel space. Left: the inputs with the trimap inset, our predicted mattes, and zoomed regions. Right: SAD, Grad and Conn on four test sets against published methods (\cref{tab:main}); each axis is the lowest plotted error divided by the method's error, so outer is lower.}
\captionsetup{hypcap=true}
\label{fig:teaser}%
\end{strip}
\begin{abstract}
Diffusion models have shown strong potential for image matting by transferring generative priors learned from large-scale pretraining. However, existing approaches largely rely on latent diffusion models, whose compressed representations and VAE-based reconstruction introduce a mismatch with the high-frequency precision required for matting. We present PixelMatte, a framework that adapts pretrained pixel-space diffusion transformers for high-fidelity image matting. We show that strong image-conditioned matting capabilities can be acquired even from a pure text-to-image model without paired image-to-image pretraining. Our approach reuses the backbone's native patch embedding to jointly process the input image, trimap, and noisy alpha matte as spatially aligned tokens, enabling direct pixel-space prediction without additional condition encoders or adapters. With pixel-accurate synthetic supervision and real-image fine-tuning, PixelMatte predicts alpha mattes in a single forward pass. Two lightweight calibration modules, LayerShift and High-Pass Refiner, further improve regional opacity accuracy and boundary fidelity. Experiments across four benchmarks and two distinct pixel diffusion backbones demonstrate substantial improvements over existing methods, reducing SAD and gradient errors by 24--44\% relative to RenderMatte on P3M-500-NP and AM-2K, while achieving competitive performance on AIM-500 and RenderMatte-2K. These results demonstrate the effectiveness of pixel-space generative priors for precision-sensitive image matting. Project page: \url{https://white-yc.github.io/pixelmatte/}.
\end{abstract}

\section{Introduction}
\label{sec:intro}
Image matting estimates the per-pixel alpha matte of an arbitrary foreground object from the observed image~\cite{li2023deep}. Unlike traditional image segmentation, which assigns discrete labels, image matting is an inherently ill-posed inverse problem due to the ambiguity in recovering alpha values from a composite image~\cite{mattingbygen,diffmatte,rendermatte}.
Trimap-guided matting addresses this ambiguity by introducing an auxiliary trimap that partitions the image into definite foreground, definite background, and an unknown region requiring exact alpha estimation.
The trimaps may be produced using segmentation models~\cite{sam,mattinganything,zhang2024click2trimap}, providing semantic and spatial guidance for image matting models to focus on recovering fine-grained details within the unknown region.
The main challenge lies in intricate structures such as hair, fur, foliage, and transparent boundaries, where accurate alpha estimation requires pixel-level precision.
Traditionally, image matting has been formulated as a supervised regression problem~\cite{dim,indexnet,gca,contextaware,fba,sim,matteformer,vitmatte,aematter,lin2025mematte}.
More recently, generative approaches have emerged that adapt pretrained diffusion models for alpha matte prediction~\cite{mattingbygen,diffmatte,diffusionmat,incontextmatting,rendermatte,omnialpha}. These methods leverage rich generative priors acquired through large-scale pretraining, which can help mitigate the effects of imperfect alpha annotations and enable generalization to diverse visual content beyond the limited distributions of existing matting datasets.

However, most existing generative approaches rely on latent diffusion models~\cite{ldm,flux}, creating an inherent mismatch with the pixel-level precision required for image matting. Matting is a high-frequency task, yet the network of a latent model operates on a spatially compressed representation in which hair-thin structures and opacity transitions are not represented explicitly~\cite{emu,pixelspacepost,pixeldit}. It predicts a compressed code and is supervised at the pixel level only through a frozen decoder trained for natural images~\cite{genpercept,layerdiffuse,drip,rechannel}. Fine details are therefore recovered indirectly, relying on the strength of the generative prior rather than on direct modeling of high-frequency alpha structure.

Recent advances in pixel-space diffusion models have made high-quality image generation directly in pixel space increasingly practical, enabled by improved prediction objectives, architectures, and training strategies~\cite{jit,pixeldit,prx,hidream,lin2026adversarial,lin2026dma}. In contrast to latent diffusion models, these approaches bypass VAE-based compression and reconstruction, preserving direct access to high-frequency image information that is only implicitly encoded in latent representations. This enables diffusion transformers to learn and optimize fine-grained visual structures under direct pixel-space supervision, making them particularly promising for image matting, where precise recovery of high-frequency details is essential.

In this paper, we present \textbf{PixelMatte}, a simple yet effective framework for high-fidelity matting built upon pretrained pixel-space diffusion transformers. We demonstrate that even a pure text-to-image model without paired image-to-image pretraining can acquire strong image-conditioned matting capabilities through supervised fine-tuning. Specifically, we directly embed the input image, trimap, and noisy alpha matte using the backbone's native patch embedding, enabling spatially aligned conditioning without additional encoders or adapters. We adapt the pretrained models using the pixel-accurate synthetic training data of RenderMatte~\cite{rendermatte}, followed by mixed training with real images. We further introduce two lightweight calibration modules, \textbf{LayerShift} and \textbf{High-Pass Refiner}, to improve opacity accuracy and boundary fidelity, respectively.

Our main contributions are summarized as follows:
\begin{itemize}
    \item We introduce \textbf{PixelMatte}, a framework that adapts pretrained pixel-space diffusion transformers to image matting by reusing the backbone's native patch embedding for spatially aligned image and trimap conditioning. It needs no VAE, condition encoder or adapter, predicts the matte in a single step, and works even for a pure text-to-image backbone (PRXPixel~\cite{prx}) without paired image-to-image pretraining.
    \item We propose two lightweight calibration modules: \textbf{LayerShift}, learned per-layer shifts that adjust regional opacity levels inside the frozen backbone, and the \textbf{High-Pass Refiner}, a small high-pass-constrained residual network that sharpens boundary transitions.
    \item With two pixel diffusion backbones, PixelMatte achieves state-of-the-art results on P3M-500-NP and AM-2K, reducing the SAD of RenderMatte~\cite{rendermatte} from 6.2 to 4.3 and from 6.5 to 4.4, and competitive results on AIM-500 and RenderMatte-2K.
\end{itemize}

\begin{figure*}[t]
  \centering
  \includegraphics[width=\textwidth]{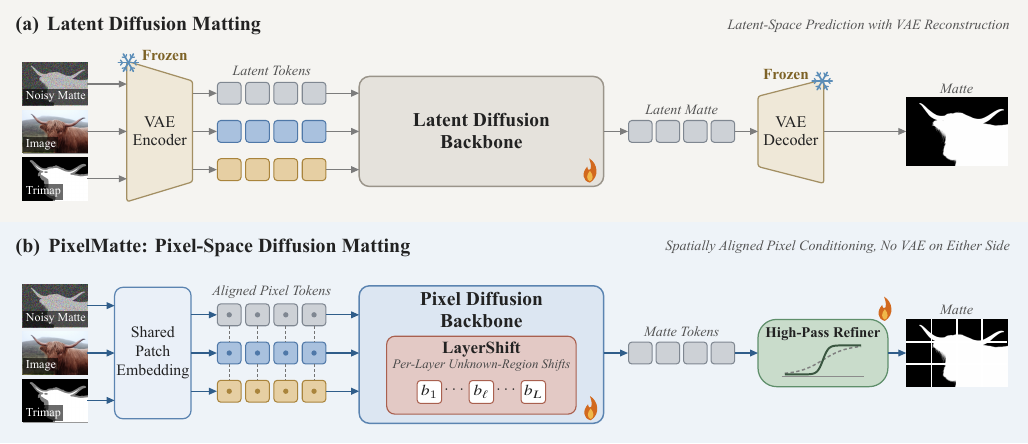}
  \caption{Overview of PixelMatte. (a) Latent diffusion matting predicts the matte as a latent code between a frozen VAE encoder and a frozen VAE decoder. (b) PixelMatte embeds the noisy matte, the image and the trimap as pixel tokens on a shared grid and predicts the matte in a single forward pass without a VAE. LayerShift acts inside the backbone, and the High-Pass Refiner refines the predicted matte.}
  \label{fig:overview}
\end{figure*}

\section{Related Work}
\label{sec:related}
\paragraph{Trimap-guided matting.}
Deep Image Matting~\cite{dim} established a learning-based framework for trimap-guided alpha estimation and introduced the Composition-1k benchmark. Subsequent methods improved alpha regression through learned feature indexing~\cite{indexnet}, guided contextual attention~\cite{gca}, attention-based representations~\cite{matteformer}, and pretrained vision transformers with detail-capture modules~\cite{vitmatte}. Real-image datasets, including P3M-10k~\cite{p3m}, AM-2K~\cite{gfm}, and AIM-500~\cite{aim}, have further advanced the evaluation of matting in diverse natural scenes. These approaches primarily learn direct alpha regression from matting-specific supervision, with high-frequency details recovered through task-specific architectures and image features. In contrast, PixelMatte explores transferring pretrained pixel-space generative priors to fine-grained alpha prediction.

\paragraph{Generative priors for image matting.}
The success of pretrained diffusion models has motivated their adaptation to dense visual prediction tasks~\cite{marigold,genpercept,lotus,e2eft,edit2perceive}. In image matting, Matting by Generation~\cite{mattingbygen} formulates alpha estimation as conditional generation using pretrained latent diffusion priors, while DRIP~\cite{drip} extends latent diffusion to joint foreground and alpha prediction with a dedicated transparency decoder. DiffMatte~\cite{diffmatte} instead introduces iterative pixel-level diffusion refinement using a lightweight decoder, and SDMatte~\cite{sdmatte} adapts Stable Diffusion for visual-prompt-guided interactive matting. More recently, RenderMatte~\cite{rendermatte} fine-tunes FLUX.1 Kontext~\cite{flux}, an image-editing model pretrained on paired image-to-image data, for trimap-guided matting, combining pixel-accurate synthetic supervision with alpha-edge losses and group-relative alignment. While these methods demonstrate the effectiveness of diffusion-based matting, approaches built upon pretrained latent image generators rely on compressed representations and decoder-based reconstruction. PixelMatte instead fine-tunes pixel-space generative backbones with the supervised recipe of RenderMatte and requires no image-editing backbone: PRXPixel is a pure text-to-image model.

\paragraph{Pixel-space diffusion transformers.}
Recent advances in pixel-space diffusion have demonstrated high-quality image generation without pretrained VAEs~\cite{pixelflow,pixnerd}. JiT~\cite{jit} revisits direct pixel prediction with large-patch transformers, while PixelDiT~\cite{pixeldit} combines patch-level generation with pixel-level detail modeling. PRXPixel~\cite{prx} and HiDream-O1-Image~\cite{hidream} extend pixel-space diffusion to large-scale image generation. Beyond generation, PixRestore~\cite{pixrestore} develops a pixel-space diffusion transformer trained from scratch for image restoration, incorporating feature-based conditioning for detail recovery. PixelControl~\cite{pixelcontrol} explores pixel-space controllable generation with additional structure-aware condition injection to improve fine-grained spatial fidelity. In contrast to these task-specific architectures, PixelMatte directly repurposes pretrained pixel-space diffusion transformers for image-conditioned alpha estimation. By reusing their native patch embedding for spatially aligned image and trimap conditioning, our framework avoids additional conditioning encoders or adapters while enabling direct pixel-space matting.

\section{Method}
\label{sec:method}

Latent diffusion matting introduces a representation mismatch between compressed latent features and the high-frequency precision required for image matting (\cref{fig:overview}a). In contrast, PixelMatte adapts pretrained pixel-space diffusion transformers to predict and supervise alpha mattes directly in pixel space, without VAE-based reconstruction (\cref{fig:overview}b). We reuse the backbone's native patch embedding to jointly process the input image, trimap, and noisy alpha matte as spatially aligned tokens (\cref{sec:cond}). The model is fine-tuned in two stages and performs alpha estimation in a single forward pass (\cref{sec:regression}). Finally, LayerShift (\cref{sec:bias}) and the High-Pass Refiner (\cref{sec:corrector}) further calibrate regional opacity levels and boundary transitions, respectively.

\subsection{Pixel-Space Matting with Aligned Conditioning}
\label{sec:cond}
Given an image $I$ and a trimap $T\in\{0,0.5,1\}^{H\times W}$ whose unknown region is $\Uband=\{p: T(p)=0.5\}$, the goal is to predict the alpha matte $\alpha\in[0,1]^{H\times W}$.
We denote by $x_0$ the matte $2\alpha-1$ replicated to three channels and by $c=(I,\bar{T})$ the condition, where $\bar{T}$ is the trimap $2T-1$ replicated to three channels.
We adopt two pretrained pixel-space rectified-flow transformers~\cite{rectifiedflow,flowmatching}, PRXPixel~\cite{prx} (PRX) and HiDream-O1-Image~\cite{hidream}.
Both operate directly on raw pixels using a linear patch embedding $E$, with patch sizes $P=16$ and $P=32$, respectively, and rotary positional embeddings~\cite{rope}.
The network $f_\theta$ predicts the clean target from its noisy counterpart:
\begin{equation}
  x_\sigma = (1-\sigma)\,x_0 + \sigma\,k\,\epsilon,\qquad \hat{x}_0 = f_\theta(x_\sigma,\sigma,c),
  \label{eq:path}
\end{equation}
where $\epsilon\sim\mathcal{N}(\mathbf{0},\mathbf{I})$, $\sigma\in(0,1]$ is the noise level and $k$ is the native noise scale of the backbone.

\paragraph{Conditioning through the patch embedding.}
A latent matting model obtains its image condition from the VAE, whose code of the image is appended to the noisy latent.
To adapt pretrained pixel-space diffusion transformers for image-conditioned matting, we reuse their native patch embedding to encode the input image and trimap alongside the noisy target matte.
Unlike latent approaches that encode image conditions through a VAE~\cite{rendermatte}, our formulation requires no separate condition encoder or adapter.
Specifically, we concatenate the three patch-token sequences:
\begin{equation}
  S = \big[\,E(x_\sigma)\;;\;E(I)+s\;;\;E(\bar{T})+s\,\big],
  \label{eq:layout}
\end{equation}
For PRX, $s$ is a learned vector initialized to zero and shared across all condition tokens; HiDream uses no such vector ($s=0$) and instead distinguishes the three segments along the temporal axis of its rotary positional encoding.
The segments share the same spatial grid, so corresponding tokens at position $(i,j)$ receive identical spatial rotary coordinates.
This alignment enables joint self-attention to directly associate target tokens with image and trimap tokens at corresponding locations.
The pretrained HiDream backbone can already exploit this spatial correspondence before matting fine-tuning, as demonstrated in our reference reconstruction experiment.
With a fixed text prompt, the target tokens are projected directly into pixel patches, enabling alpha prediction and supervision without an intermediate VAE decoder.

\subsection{Pixel-Accurate Training and Inference}
\label{sec:regression}
\paragraph{Training objective.}
We adopt the supervised objective of RenderMatte~\cite{rendermatte}, a flow-matching term in the unknown region plus its alpha-edge loss $\LAE$ on the predicted matte,
\begin{equation}
  \begin{aligned}
  \mathcal{L} = {}& \mathbb{E}_{(x_0,c),\sigma,\epsilon}\Big[\tfrac{w(\sigma)}{\sigma^{2}}\,\big\|f_\theta(x_\sigma,\sigma,c)-x_0\big\|_{\Uband}^2 \\
  & + \lambda\,\LAE\!\big(\mean_{\mathrm{ch}} f_\theta(x_\sigma,\sigma,c),\,2\alpha-1;\,\Uband\big)\Big],
  \end{aligned}
  \label{eq:loss}
\end{equation}
where $\|\cdot\|_{\Uband}^2$ is the squared error restricted to $\Uband$ and averaged over the image, $\mean_{\mathrm{ch}}$ is the channel mean, $w(\sigma)$ is a bell-shaped weight with $w(1)=0$, $\lambda$ is RenderMatte's adaptive balancing weight, and $\LAE$ sums $\ell_1$, SAD, MSE, gradient and Laplacian-pyramid terms.
The first term is equivalent to a weighted velocity-based flow-matching loss, since the velocity prediction error equals the clean-target prediction error divided by $\sigma$, yielding the factor $1/\sigma^{2}$.
In the latent baseline $\LAE$ is applied to a VAE-decoded estimate; here it supervises the prediction directly at pixel level, with no decoder in between.

\paragraph{Two-stage fine-tuning.}
Both backbones are fine-tuned with all transformer weights in two stages. The synthetic stage uses the synthetic composite training set of RenderMatte~\cite{rendermatte}, which provides accurate alpha supervision. To bridge the synthetic-to-real domain gap, the shorter real-data stage retains synthetic composites for 75\% of the samples and draws the remaining 25\% from the P3M-10k~\cite{p3m} and AM-2K~\cite{gfm} training splits.

\paragraph{One-step inference.}
At test time we set $\sigma=1$, draw one noise field from a fixed seed, and obtain the matte from a single forward pass,
\begin{equation}
  \ahat = \clip\!\big(\tfrac12\big(\mean_{\mathrm{ch}} f_\theta(k\epsilon,\,1,\,c)+1\big),\,0,\,1\big),
  \label{eq:onestep}
\end{equation}
and we set $\ahat(p)=T(p)$ for $p\notin\Uband$.
At $\sigma=1$ the input is independent of $x_0$, so the network acts as a regressor of the matte given the image and the trimap, with the noise field as its only other input.

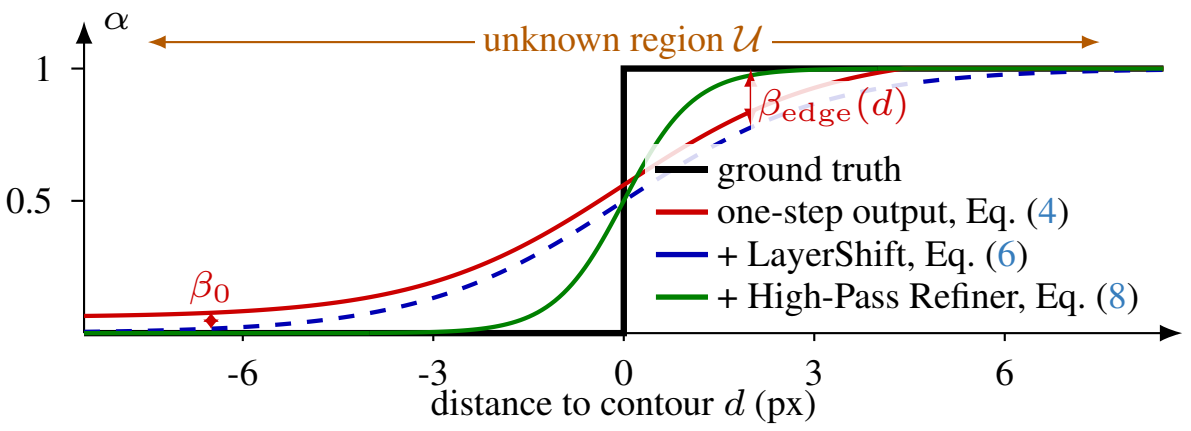
\begin{figure}[t]
  \centering
  \begin{tikzpicture}[font=\scriptsize, x=0.42cm, y=1.75cm]
    \draw[-{Latex[length=1.5mm]}] (-8.5,0) -- (8.8,0);
    \node at (0,-0.275) {distance to contour $d$ (px)};
    \draw[-{Latex[length=1.5mm]}] (-8.5,0) -- (-8.5,1.18) node[right] {$\alpha$};
    \foreach \x in {-6,-3,0,3,6} {\draw (\x,0) -- (\x,-0.03) node[below] {\x};}
    \draw (-8.5,1) -- (-8.7,1) node[left] {1};
    \draw (-8.5,0.5) -- (-8.7,0.5) node[left] {0.5};
    \draw[{Latex[length=1.2mm]}-{Latex[length=1.2mm]}, orange!70!black] (-7.5,1.1) -- (7.5,1.1) node[midway, fill=white, inner sep=1.5pt] {unknown region $\Uband$};
    \draw[very thick] (-8.5,0) -- (0,0) -- (0,1) -- (8.5,1);
    \draw[thick, red!80!black, domain=-8.5:8.5, samples=80, smooth] plot (\x, {min(1, 0.06 + 1/(1+exp(-\x/1.6)))});
    \draw[thick, dashed, blue!70!black, domain=-8.5:8.5, samples=80, smooth] plot (\x, {1/(1+exp(-\x/1.6))});
    \draw[thick, green!50!black] (-8.5,0) -- (-4,0);
    \draw[thick, green!50!black, domain=-4:4, samples=60, smooth] plot (\x, {1/(1+exp(-\x/0.55))});
    \draw[thick, green!50!black] (4,1) -- (8.5,1);
    \draw[{Latex[length=0.6mm,width=1.1mm]}-{Latex[length=0.6mm,width=1.1mm]}, red!80!black] (-6.5,0.017) -- (-6.5,0.077) node[pos=1, above, inner sep=1.5pt] {$\beta_0$};
    \draw[{Latex[length=1.2mm]}-{Latex[length=1.2mm]}, red!80!black] (2.0,0.78) -- (2.0,1.0) node[pos=0.35, right, fill=white, fill opacity=0.85, text opacity=1, inner sep=1pt] {$\beta_{\mathrm{edge}}(d)$};
    \node[anchor=south east, align=left, inner sep=2pt, fill=white, fill opacity=0.85, text opacity=1] at (8.3,0.02) {
      \textcolor{black}{\rule[0.5ex]{1.4em}{1.2pt}} ground truth\\
      \textcolor{red!80!black}{\rule[0.5ex]{1.4em}{1pt}} one-step output, \cref{eq:onestep}\\
      \textcolor{blue!70!black}{\rule[0.5ex]{1.4em}{1pt}} + LayerShift, \cref{eq:bias}\\
      \textcolor{green!50!black}{\rule[0.5ex]{1.4em}{1pt}} + High-Pass Refiner, \cref{eq:corrector}};
  \end{tikzpicture}
  \caption{Schematic illustration of regional alpha-level offsets and boundary smoothing in one-step matting. LayerShift adjusts the level offset $\beta_0$, and the High-Pass Refiner sharpens the over-wide transition $\beta_{\mathrm{edge}}(d)$. The profiles are illustrative, not measured.}
  \label{fig:bias}
\end{figure}

\paragraph{Systematic errors of the one-step output.}
The one-step output shows two recurring error patterns (\cref{fig:bias}): over-wide hard edges and regional offsets of the alpha value.
The former may arise from the boundary uncertainty of a regressor and from labels softened by resampling; a possible factor for the latter is the weak supervision at $\sigma=1$, where the flow-matching weight vanishes.
To motivate the two calibration modules, we write the error $e(p)=\ahat(p)-\alpha(p)$ near a hard edge schematically as
\begin{equation}
  e(p) \approx \underbrace{\beta_0\vphantom{\big(}}_{\text{level}} + \underbrace{\beta_{\mathrm{edge}}\big(d(p)\big)}_{\text{edge profile}} + r(p),
  \label{eq:decomp}
\end{equation}
where $d(p)$ is the signed distance to the ground-truth edge, $\beta_0$ a regional offset of the alpha level, $\beta_{\mathrm{edge}}$ the error of the edge profile, and $r$ the remaining image-dependent error.
The decomposition is illustrative rather than an assumption of strictly separable errors, and motivates one module for each term.

\begin{table*}[t]
  \centering
  \scriptsize
  \setlength{\tabcolsep}{1.8pt}
  \caption{Quantitative comparison on P3M-500-NP, AM-2K, AIM-500 and RenderMatte-2K (RM-2K). Lower is better for all metrics; MSE is reported in units of $10^{-3}$; the best results are in bold. $\dagger$: native interaction format instead of a trimap.}
  \label{tab:main}
  \resizebox{\textwidth}{!}{%
  \begin{tabular}{l cccc cccc cccc cccc}
    \toprule
    & \multicolumn{4}{c}{P3M-500-NP} & \multicolumn{4}{c}{AM-2K} & \multicolumn{4}{c}{AIM-500} & \multicolumn{4}{c}{RM-2K} \\
    \cmidrule(lr){2-5}\cmidrule(lr){6-9}\cmidrule(lr){10-13}\cmidrule(lr){14-17}
    Method & SAD & MSE & Grad & Conn & SAD & MSE & Grad & Conn & SAD & MSE & Grad & Conn & SAD & MSE & Grad & Conn \\
    \midrule
    DIM~\cite{dim} & 24.5 & 7.0 & 29.2 & 20.6 & 40.5 & 13.0 & 32.7 & 33.3 & 53.1 & 17.0 & 42.3 & 37.9 & 34.0 & 19.0 & 39.1 & 22.1 \\
    GCA~\cite{gca} & 6.2 & 1.0 & 9.8 & 4.5 & 6.3 & 0.9 & 5.5 & 4.8 & 23.4 & 7.6 & 15.9 & 12.1 & 8.4 & 4.0 & 12.1 & 5.0 \\
    IndexNet~\cite{indexnet} & 7.2 & 1.0 & 9.5 & 5.5 & 9.2 & 2.0 & 8.1 & 7.3 & 25.8 & 7.0 & 16.3 & 17.1 & 11.4 & 5.0 & 16.0 & 7.2 \\
    MatteFormer~\cite{matteformer} & 7.9 & 2.0 & 11.8 & 5.7 & 8.5 & 2.0 & 7.2 & 6.3 & 27.0 & 9.0 & 20.3 & 15.1 & 8.6 & 4.0 & 13.9 & 4.7 \\
    ViTMatte~\cite{vitmatte} & 7.1 & 1.0 & 10.6 & 5.0 & 7.7 & 1.0 & 6.2 & 5.5 & 17.8 & 4.0 & 13.7 & 11.0 & 4.5 & 1.0 & 4.8 & 3.0 \\
    DiffMatte~\cite{diffmatte} & 6.6 & 1.0 & 10.0 & 4.8 & 6.8 & 1.0 & 5.7 & 5.0 & 17.2 & 4.0 & 13.5 & 11.6 & 4.3 & 1.0 & 4.4 & 3.0 \\
    SDMatte~\cite{sdmatte}$^\dagger$ & 32.0 & 13.0 & 20.4 & 20.8 & 17.5 & 6.0 & 13.2 & 10.9 & 31.8 & 11.0 & 26.8 & 17.5 & 12.1 & 8.0 & 20.4 & 3.6 \\
    Edit2Perceive~\cite{edit2perceive}$^\dagger$ & 19.4 & 3.0 & 13.2 & 10.2 & 20.4 & 4.0 & 9.6 & 9.9 & 29.1 & 6.0 & 18.2 & 15.7 & 11.5 & 5.0 & 9.2 & 5.0 \\
    RenderMatte~\cite{rendermatte} & 6.2 & 1.0 & 8.2 & 4.8 & 6.5 & 1.0 & 5.5 & 5.2 & 13.6 & \textbf{2.0} & 11.2 & 9.8 & 3.6 & 1.0 & 3.0 & 2.6 \\
    \midrule
    PixelMatte (PRX-7B) & 4.7 & 0.6 & 5.6 & 3.7 & 4.6 & 0.5 & 3.4 & 3.9 & 12.7 & 2.2 & 8.5 & \textbf{9.3} & 3.5 & 0.9 & 3.2 & 2.4 \\
    PixelMatte (HiDream-8B) & \textbf{4.3} & \textbf{0.5} & \textbf{4.6} & \textbf{3.5} & \textbf{4.4} & \textbf{0.4} & \textbf{3.1} & \textbf{3.8} & \textbf{12.6} & 2.1 & \textbf{7.7} & 9.5 & \textbf{3.3} & \textbf{0.8} & \textbf{2.8} & \textbf{2.3} \\
    \bottomrule
  \end{tabular}}
\end{table*}

\subsection{LayerShift for Alpha-Level Calibration}
\label{sec:bias}
LayerShift is designed to calibrate regional alpha-level offsets through lightweight modifications to the frozen backbone.
Let $h_{\ell,n}\in\mathbb{R}^{C}$ be the hidden state of target token $n$ in layer $\ell$ after the second normalization and before the MLP, and let $u_n\in\{0,1\}$ indicate whether its patch overlaps the unknown region.
LayerShift adds one learned vector $b_\ell\in\mathbb{R}^{C}$ per layer to these target tokens,
\begin{equation}
  h_{\ell,n} \;\leftarrow\; h_{\ell,n} + u_n\, b_\ell,\qquad \ell=1,\dots,L.
  \label{eq:bias}
\end{equation}
The vectors are trained for 300 steps at $\sigma=1$ with $\LAE$ while the backbone stays frozen; for HiDream this is $L\times C = 36\times4096 = 147{,}456$ parameters.
Shared across images and spatial locations, these vectors provide a compact, input-independent parameterization for calibrating regional alpha levels; their effect on the output still depends on the activations of the backbone.

\subsection{High-Pass Refiner for Boundary Calibration}
\label{sec:corrector}
The High-Pass Refiner targets the edge-profile term with a small residual network whose output passes through a high-pass operator.
Correcting an overly smooth transition often involves changes of opposite sign on the two sides of an edge, rather than a uniform shift of the alpha values.
We encourage such corrections through the architecture rather than through the loss: for a residual field $D$,
\begin{equation}
  \Php^{\Uband}(D) \;=\; m_{\Uband}\odot\Big(D - \frac{G_{\sigma_h}\ast(m_{\Uband}\odot D)}{\max(G_{\sigma_h}\ast m_{\Uband},\,\eta)}\Big),
  \label{eq:php_band}
\end{equation}
where $m_{\Uband}$ is the binary mask of the unknown region, $G_{\sigma_h}$ is a Gaussian kernel with standard deviation $\sigma_h=24$\,px, and $\eta=10^{-3}$.
By subtracting the mask-normalized local mean of $D$, the operator suppresses slowly varying corrections and retains localized ones, which encourages boundary refinement rather than regional shifts of the alpha level.
A small CNN $g_\phi$ (8 dilated residual blocks, 96 channels, 1.3M parameters) takes the one-step matte, the trimap and the image and predicts $D=\tfrac12\tanh(g_\phi(\ahat,T,I))$; the refined matte is
\begin{equation}
  \acorr \;=\; \clip\!\big(\ahat + \Php^{\Uband}(D),\,0,\,1\big).
  \label{eq:corrector}
\end{equation}
$g_\phi$ is trained with $\LAE$ on cached one-step predictions of the frozen backbone without LayerShift, and is applied at inference to the output with LayerShift.
Without the subtraction, the SAD is lower on the synthetic benchmark but higher on two real test sets.

\section{Experiments}
\label{sec:experiments}

\subsection{Experimental Setup}
\label{sec:setup}
\paragraph{Implementation details.}
We evaluate PixelMatte with two pretrained pixel-space diffusion transformers, PRX-7B (PRXPixel~\cite{prx}) and HiDream-O1-Image 8B~\cite{hidream}.
Both follow the fine-tuning procedure of \cref{sec:regression}; PRX operates at $1024^2$, while HiDream is adapted to $2048^2$ for its final training stages and inference.
PixelMatte (PRX-7B) and PixelMatte (HiDream-8B) denote the complete models with LayerShift and the High-Pass Refiner.
Inference is a single forward pass without test-time augmentation, and we report results averaged over two independently evaluated noise seeds.
On a single H200 GPU, PRX and HiDream take 0.46\,s at $1024^2$ and 0.64\,s at $2048^2$, respectively, compared with 0.91\,s for RenderMatte at $1024^2$.

\paragraph{Datasets, metrics and baselines.}
We evaluate on P3M-500-NP~\cite{p3m}, AM-2K~\cite{gfm}, AIM-500~\cite{aim} and RenderMatte-2K (RM-2K)~\cite{rendermatte}, using their released trimaps.
We report SAD, MSE ($\times10^{3}$), Grad and Conn.
In \cref{tab:main}, Conn follows the implementation released with RenderMatte for consistency with the published results; the remaining tables use the definition of DIM~\cite{dim}.
We compare against representative supervised and diffusion-based matting methods using the published results collected by RenderMatte~\cite{rendermatte}.

\subsection{Comparison Experiments}
\label{sec:main_results}

\paragraph{Quantitative comparison.}
\Cref{tab:main} compares PixelMatte with existing matting methods on the four benchmarks.
On P3M-500-NP and AM-2K, which are in-domain for our real-data stage, PixelMatte (HiDream-8B) reduces SAD and Grad by 30--44\% relative to RenderMatte, and PixelMatte (PRX-7B), which runs at the same $1024^2$ resolution as RenderMatte, by 24--38\%.
These gains are already evident in the one-step predictions before calibration, and the HiDream model trained only in the synthetic stage already achieves a lower Grad than RenderMatte on all three real-image benchmarks.
On AIM-500, both backbones reduce Grad by 24--31\%, indicating improved accuracy in fine spatial transitions.
On the synthetic RM-2K benchmark, PixelMatte performs comparably to RenderMatte.
These results show that pixel-space diffusion is an effective alternative to latent generative matting, particularly on fine structures in real images.

\begin{figure*}[p]
  \begin{minipage}[t][0.995\textheight][t]{\textwidth}
  \centering
  \includegraphics[width=\textwidth]{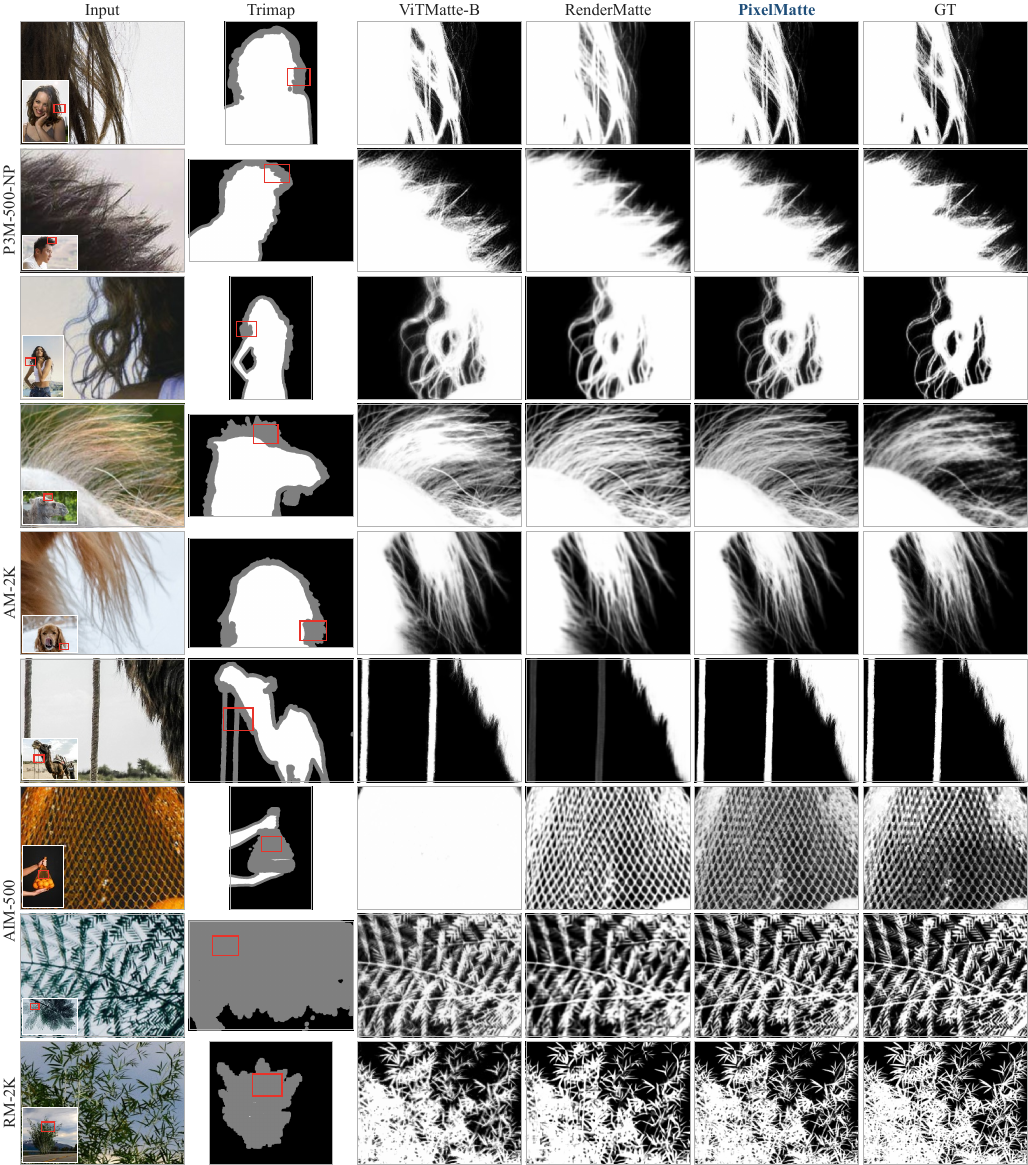}
  \caption{Qualitative comparison on P3M-500-NP, AM-2K, AIM-500 and RM-2K. From left to right: input image, trimap, ViTMatte-B, RenderMatte, PixelMatte and the ground-truth alpha matte. Each row shows a zoomed region.}
  \label{fig:qual}
  \end{minipage}
\end{figure*}

\paragraph{Qualitative comparison.}
\Cref{fig:qual} presents visual comparisons on the four benchmarks, with zoomed regions selected independently of the methods.
PixelMatte recovers fine alpha structures such as hair strands and animal fur, which complements its lower gradient error.

\subsection{Ablation Studies}
\label{sec:calib}
\Cref{tab:ablation} ablates the two modules on RM-2K for both backbones, and \cref{tab:cf} ablates the image input of the High-Pass Refiner.

\begin{table}[t]
  \centering
  \scriptsize\setlength{\tabcolsep}{4pt}
  \caption{Ablation of LayerShift and the High-Pass Refiner on RM-2K. Lower is better for all metrics; the best results per backbone are in bold.}
  \label{tab:ablation}
  \resizebox{\columnwidth}{!}{%
  \begin{tabular}{l cccc}
    \toprule
    Configuration & SAD & MSE & Grad & Conn \\
    \midrule
    \multicolumn{5}{l}{\emph{HiDream-8B}} \\
    one-step output, \cref{eq:onestep} & 3.74 & 0.84 & 2.94 & 3.42 \\
    + High-Pass Refiner, \cref{eq:corrector} & 3.38 & 0.81 & 2.83 & 3.12 \\
    + LayerShift, \cref{eq:bias} & 3.66 & 0.82 & 2.90 & 3.33 \\
    + LayerShift + High-Pass Refiner & \textbf{3.31} & \textbf{0.80} & \textbf{2.81} & \textbf{3.04} \\
    \midrule
    \multicolumn{5}{l}{\emph{PRX-7B}} \\
    one-step output, \cref{eq:onestep} & 3.60 & 0.87 & 3.30 & 3.28 \\
    + LayerShift, \cref{eq:bias} & 3.59 & 0.88 & 3.31 & 3.26 \\
    + LayerShift + High-Pass Refiner & \textbf{3.46} & \textbf{0.86} & \textbf{3.25} & \textbf{3.18} \\
    \bottomrule
  \end{tabular}}
\end{table}

\begin{table}[t]
  \centering
  \scriptsize\setlength{\tabcolsep}{2pt}
  \caption{Dependence of the High-Pass Refiner on its RGB input (HiDream-8B with LayerShift, SAD, one noise seed). ``Base'': no refinement. The image is replaced at inference; the last row retrains the refiner without RGB. Lower is better; best in bold.}
  \label{tab:cf}
  \resizebox{\columnwidth}{!}{
  \begin{tabular}{l cccc}
    \toprule
    Refiner input & P3M-500-NP & AM-2K & AIM-500 & RM-2K \\
    \midrule
    none (base) & 4.52 & 4.68 & 12.39 & 3.67 \\
    true image & \textbf{4.35} & \textbf{4.45} & \textbf{12.12} & 3.32 \\
    blur $\sigma=4$ px & 4.40 & 4.54 & 12.24 & 3.32 \\
    another image & 4.40 & 4.53 & 12.23 & 3.34 \\
    constant & 4.44 & 4.57 & 12.26 & 3.33 \\
    retrained without image & 4.40 & 4.50 & 12.21 & \textbf{3.29} \\
    \bottomrule
  \end{tabular}}
\end{table}

\begin{table}[t]
  \centering
  \scriptsize\setlength{\tabcolsep}{3.5pt}
  \caption{SAD on RM-2K by ground-truth alpha value and contour distance (HiDream-8B). Values are additive contributions to the total, up to rounding. Lower is better.}
  \label{tab:where}
  \resizebox{\columnwidth}{!}{%
  \begin{tabular}{l ccc}
    \toprule
    Pixel subset & one-step & + LayerShift & PixelMatte \\
    \midrule
    all pixels & 3.76 & 3.67 & 3.32 \\
    \midrule
    GT fractional ($0<\alpha<1$) & 3.01 & 2.95 & 2.94 \\
    GT saturated ($\alpha\in\{0,1\}$) & 0.75 & 0.72 & 0.38 \\
    \quad within 3 px of contour & 0.56 & 0.54 & 0.23 \\
    \quad 3--8 px from contour & 0.09 & 0.09 & 0.06 \\
    \quad beyond 8 px & 0.10 & 0.10 & 0.08 \\
    \bottomrule
  \end{tabular}}
\end{table}

\begin{figure*}[t]
  \centering
  \includegraphics[width=\textwidth]{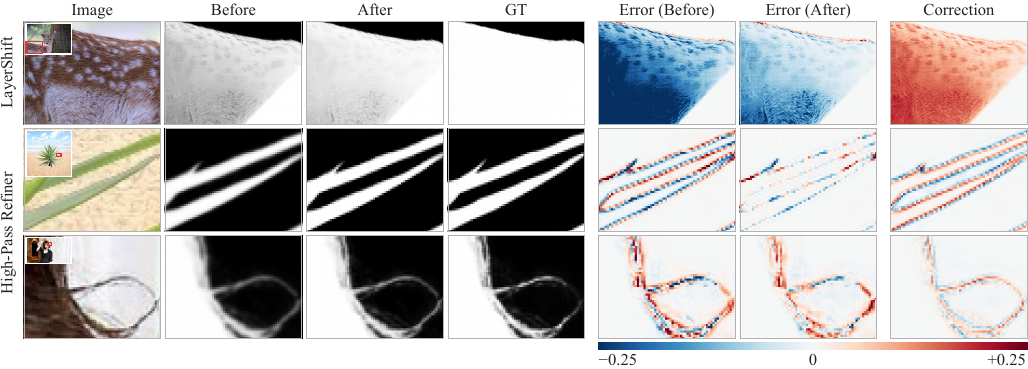}
  \caption{Visualization of opacity and boundary calibration. Each row compares the alpha mattes before and after a calibration module against the ground truth. Error maps show prediction minus ground truth, while correction maps show the matte after minus before, using a shared color scale. LayerShift adjusts regional opacity levels (top), whereas the High-Pass Refiner concentrates its corrections around fine boundaries (middle and bottom).}
  \label{fig:calib}
\end{figure*}

\paragraph{Component ablation.}
On HiDream-8B, LayerShift reduces the RM-2K SAD from 3.74 to 3.66 with only 147K trainable parameters, and the High-Pass Refiner alone reaches 3.38.
Combining both modules reduces SAD to 3.31 and improves all four metrics (\cref{tab:ablation}).
The refiner maintains a similar relative SAD improvement when combined with LayerShift, indicating that boundary refinement remains effective after opacity calibration.
The combined calibration also improves PRX-7B on all four metrics, reducing its SAD from 3.60 to 3.46.
On the real-image benchmarks, the High-Pass Refiner further reduces the SAD of the LayerShift-calibrated HiDream predictions by 3.7\%, 5.2\% and 2.1\% on P3M-500-NP, AM-2K and AIM-500, respectively.

\paragraph{Effect of the image input.}
\label{sec:counterfactual}
We examine how the High-Pass Refiner uses RGB information by replacing its image input at inference or retraining it without RGB (\cref{tab:cf}).
On RM-2K, replacing the image with an unrelated one preserves about 95\% of the refiner's SAD improvement, and the refiner retrained without RGB performs comparably (3.29 versus 3.32).
On the three real-image benchmarks, the RGB-free refiner retains 68--75\% of the improvement of the full input.
The refiner thus calibrates boundaries already predicted by the backbone rather than re-estimating the matte from the image, with RGB providing additional benefits on real images, so the fine structures of the final matte originate from the pixel-space backbone.

\subsection{Analysis and Discussion}
\label{sec:where}
\paragraph{Effects of calibration.}
\Cref{tab:where} decomposes SAD by the ground-truth alpha value and the distance to the contour.
LayerShift provides modest improvements across all pixel subsets, consistent with its role in regional opacity calibration.
In contrast, the High-Pass Refiner achieves its largest gains near hard boundaries: on saturated pixels within 3\,px of the contour, the SAD contribution decreases from 0.54 after LayerShift to 0.23 after refinement.
\Cref{fig:calib} further distinguishes the two behaviors: LayerShift adjusts opacity over extended regions, whereas the High-Pass Refiner introduces localized corrections along object contours.

\paragraph{Application to a latent backbone.}
\label{sec:specific}
We further apply the same refiner architecture to RenderMatte, a latent diffusion matting model; each refiner is trained on the predictions of its own backbone using only RenderMatte's synthetic composite training set (\cref{tab:specific}).
Both models benefit from refinement on all four benchmarks.
On RM-2K, the refiner reduces the SAD of HiDream from 3.74 to 3.37 and that of RenderMatte from 3.33 to 3.28, which shows that learned boundary calibration applies to different generative matting backbones.

\begin{table}[t]
  \centering
  \scriptsize\setlength{\tabcolsep}{2pt}
  \caption{High-Pass Refiner on a pixel-space and a latent one-step model (SAD); each refiner is trained on its own backbone's predictions with RenderMatte's synthetic training set. Lower is better.}
  \label{tab:specific}
  \resizebox{\columnwidth}{!}{
  \begin{tabular}{l cccc}
    \toprule
    Model & P3M-500-NP & AM-2K & AIM-500 & RM-2K \\
    \midrule
    HiDream-8B, one-step & 4.52 & 4.87 & 13.94 & 3.74 \\
    \quad + High-Pass Refiner & 4.39 & 4.74 & 13.75 & 3.37 \\
    RenderMatte, one-step & 5.87 & 5.94 & 13.21 & 3.33 \\
    \quad + High-Pass Refiner & 5.72 & 5.80 & 12.85 & 3.28 \\
    \bottomrule
  \end{tabular}}
\end{table}

\paragraph{Limitations.}
\label{sec:limitations}
Although direct pixel-space prediction removes the compressed latent representation and its decoder, it does not remove the limitations of generative pretraining or of matting supervision.
PixelMatte inherits visual priors learned primarily for image synthesis, which may not fully capture the opacity characteristics of challenging materials: we observe residual low-amplitude alpha in saturated regions on RM-2K and noise-seed sensitivity for some transparent objects on AIM-500.
Our calibration modules, which are trained separately for each backbone, improve regional opacity and boundary transitions but do not resolve every such error.

\section{Conclusion}
\label{sec:conclusion}
We presented PixelMatte, demonstrating that pretrained pixel-space diffusion models can be effectively adapted to high-fidelity image matting through simple spatially aligned conditioning and task-specific supervision. By directly predicting alpha mattes in pixel space, our approach avoids VAE-based reconstruction while retaining the visual priors acquired during generative pretraining. Lightweight opacity and boundary calibration modules further improve prediction fidelity, yielding strong results across two distinct diffusion backbones and multiple matting benchmarks. Our findings suggest that pixel-space generative foundation models offer a promising alternative to latent diffusion for precision-sensitive dense prediction tasks, extending their potential beyond image synthesis.

\section*{Acknowledgments}
This work used Anvil AI~\cite{anvil} at Purdue University through allocation ELE260077 from the Advanced Cyberinfrastructure Coordination Ecosystem: Services \& Support (ACCESS) program~\cite{access}, which is supported by U.S. National Science Foundation grants \#2138259, \#2138286, \#2138307, \#2137603, and \#2138296.

{
    \small
    \bibliographystyle{ieeenat_fullname}
    \bibliography{main}
}

\end{document}